\documentclass{article}
\usepackage[top=3cm,bottom=3cm,left=3.9cm,right=3.9cm]{geometry}
\usepackage{amsmath}
\usepackage[round]{natbib}
\usepackage{tipa}
\usepackage{color}

\DeclareMathOperator*{\argmax}{arg\,max}

\newcommand{\len}{s}

\begin{document}

\title{A Probabilistic Approach to\\Pronunciation by Analogy}
\author{Janne V. Kujala\footnote{Address: Department of Mathematical
    Information Technology, University of Jyv{\"a}skyl{\"a}, Finland.
    Email address: \texttt{jvk@iki.fi}} \and Aleksi
  Keurulainen\footnote{Address: Agora Center, University of
    Jyv{\"a}skyl{\"a}, Finland}} \date{September 21, 2011}

\maketitle

\begin{abstract}
  The relationship between written and spoken words is convoluted in
  languages with a deep orthography such as English and therefore it
  is difficult to devise explicit rules for generating the
  pronunciations for unseen words. Pronunciation by analogy (PbA) is a
  data-driven method of constructing pronunciations for novel words
  from concatenated segments of known words and their
  pronunciations. PbA performs relatively well with English and
  outperforms several other proposed methods. However, the best
  published word accuracy of 65.5\% (for the 20,000 word NETtalk
  corpus) suggests there is much room for improvement in it.

  Previous PbA algorithms have used several different scoring
  strategies such as the product of the frequencies of the component
  pronunciations of the segments, or the number of different
  segmentations that yield the same pronunciation, and different
  combinations of these methods, to evaluate the candidate
  pronunciations.  In this article, we instead propose to use a
  probabilistically justified scoring rule.  We show that this
  principled approach alone yields better accuracy (66.21\% for the
  NETtalk corpus) than any previously published PbA algorithm.
  Furthermore, combined with certain ad hoc modifications motivated by
  earlier algorithms, the performance climbs up to 66.6\%, and further
  improvements are possible by combining this method with other
  methods.
\end{abstract}

\section{Introduction}

Pronunciation by analogy (PbA) was originally proposed by
\cite{Glushko1979} as a psychological model of how humans
pronounce words.  He suggested that instead of retrieving a single
pronunciation from memory for known words and using abstract
spelling-to-sound rules for pseudowords, humans pronounce both words
and pseudowords using a similar process based on analogy to similar
words.

The first concrete definition of PbA was given by the PRONOUNCE
computer program of \cite{DedinaNusbaum1991}.  The program uses four
components: 1) a \emph{lexical database} to store the pronunciations
of known words, 2) a \emph{matcher} to list segments of the input word
that match segments of known words in the lexical database, a 3)
\emph{pronunciation lattice} to arrange the segments and their
pronunciations within the lexical database into a graph-like
data structure representing possible pronunciations of the input word,
and 4) a \emph{decision function} to select the best pronunciation for
the input word among the different candidate pronunciations.  The same
basic structure has been used in PbA implementations since then, but
the different components have gone through various improvements.

PbA is currently one of the most accurate methods for automatically
generating the pronunciations of unseen words.  However, all PbA
algorithms yield several candidate pronunciations and the selection of
the best one is generally based on ad hoc methods.  Indeed,
\cite{DamperMarchand2006} state that little or no theory exists to
guide the selection and the desire for a more probabilistic
formulation was already apparent in \cite{DamperEastmond1997} and
\cite{DamperMarchand1998}.  In the lack of a theoretically
preferable approach, an important idea leading to the current state of
the art is the multi-strategy approach of
\cite{DamperMarchand2000} which combines several evaluation
methods in the decision function to squeeze out the best accuracy
scores \citep{DamperMarchand2006,Polyakova2008,Polyakova2009}.

In this article, we instead propose to use a probabilistically
justified evaluation function for choosing the best candidate
pronunciation.  We show that a single, principled method can yield
better results than any of the previously published algorithms.
Furthermore, our analysis of the best previously published single
strategy, the PFSP method of \cite{Polyakova2008}, suggests the
use of what we call ``magic root'' in this article, a certain power
function applied in the evaluation function.  Although it is difficult
to justify this theoretically, we show that it can be used to further
improve the accuracy of our otherwise principled algorithm.

In the rest of this section, we describe the four basic components of
a PbA system in more detail, review the relevant literature, and
present the results of our reimplementation of the best published PbA
algorithms.  In Section~\ref{sec:probapproach}, we present and
evaluate our simple probabilistically justified evaluation method
using non-overlapping segments; in Section~\ref{sec:proboverlapping},
we generalize the method for overlapping segments and consider the
``magic root'' mentioned above for improving performance.  Then, we
conclude.

\subsection{Lexical Database}

The lexical database is assumed to consist of a list of
\emph{aligned} word-pronunciation pairs $(x,y)=(x_1\dots x_l,y_1\dots
y_l)$ so that there is a one-to-one correspondence of the letters (or
graphemes) $x_i$ to the phonemes $y_i$.  Thus, the English word
\emph{that} should ideally be aligned as $(x_1,x_2,x_3)=($\emph{th},
\emph{a}, \emph{t}$)$ and $(y_1,y_2,y_3)=($\textipa{D}, \textipa{\ae},
\textipa{t}$)$, where we have used the IPA phonetic symbols for the
phonemes.  To facilitate direct comparison to other PbA work, we have
for now used the NETtalk corpus which inserts null phonemes
\textipa{/-/} (e.g., \emph{th} is transcribed as \textipa{/D-/}) and
represents certain combinations as single phonemes (e.g.,
\textipa{/ks/} replaced by \textipa{/}X\textipa{/}) so as to yield
single-letter-to-phoneme alignment.  However, all the algorithms can
be naturally generalized to use many-to-many alignment.

\subsection{Matcher}

\mbox{}\cite{DedinaNusbaum1991} defined the matcher as follows.
Given an input word, all the words in the lexical database are
iterated through.  For each lexical entry, the matching procedure
starts with the input word and the lexical entry left-aligned and then
slides the shorter word to the right one letter at a time until the
two words are right aligned.  At each step, all substrings of letter
positions where the two strings match are extracted together with the
corresponding pronunciation of the lexical entry and stored in the
pronunciation lattice (described below).

\cite{DamperMarchand1998} generalized this procedure to so called
\emph{full matching}, where the matching is done over the range of
all possible overlaps of the two strings so that all common substrings
are extracted regardless of their positions within the words.  This
matching procedure is in fact equivalent to the approach described by
\cite{SullivanDamper1993}, where the frequencies of all substrings
of the lexicon together with their pronunciations are first enumerated
and stored in a dictionary, and then, for each input word, the
frequencies of its substrings are looked up in the precomputed
dictionary to form the pronunciation lattice.

\cite{DedinaNusbaum1991} and \cite{DamperMarchand1998} match
the beginning of a word only to beginnings of other words and the end
of a word only to the ends of other words.
\cite{SullivanDamper1993} investigate the effect of this
restriction.  In their approach, this amounts to storing a word as
\emph{\#word\#}, padded with silent beginning and end characters
\emph{\#} that match the beginnings and ends of other words.  In most
cases, they obtained the best results by using the word boundary
characters (see the reference for details) and this result was also
confirmed by our experiments.

\subsection{Pronunciation Lattice}\label{sec:pronlattice}

In \cite{DedinaNusbaum1991} and \cite{DamperMarchand1998}, the
pronunciation of a word is always constructed from segments that
overlap by one letter and have matching pronunciations for the overlap
letter.  The nodes of the pronunciation lattice represent the overlap
letters, identified by their pronunciation and position in the target
word.  In addition, special Start and End nodes represent the
beginning and end of the word (corresponding to the silent word
boundary null phonemes).  Each arc represents a matched segment of the
target word between and including the overlap letters.  The arc is
labeled by the pronunciation of the segment in the matched word and
the frequency of this pronunciation of the segment in all matched
words.  Each path from Start to End represents one possible
pronunciation for the word.

\cite{Yvon1996} generalized the one character overlap to an
unbounded overlap, with the nodes of the pronunciation lattice given
by all matched substrings of the input word identified by their letter
positions and pronunciations in matched words.  This results in
pronunciations given by several multiply overlapping chunks, for
example, a five letter word could be segmented into three segments at
letter positions 1--3, 2--4, and 3--5 with all three segments
overlapping the letter position 3.

\cite{SullivanDamper1993} define the pronunciation lattice
differently, with the nodes representing the junctures between
letters.  In this case, any path from Start to End represents one
possible pronunciation consisting of non-overlapping segments.  They
label the arcs with probability-like ``preference values'' instead of
plain frequencies.

\subsection{Decision Function}

In their original article, \cite{DedinaNusbaum1991} use a scoring
heuristic that chooses the shortest path (i.e., the pronunciation with
the smallest number of segments) and in case of a tie, chooses the
path with the greatest sum of the arc frequencies.

\cite{DamperEastmond1997} also use shortest paths as the primary
heuristic but argue that a product of the frequencies is intuitively
better as it is closer to a probabilistic formulation.  As a secondary
heuristic they use the maximum product of arc frequencies.
\cite{SullivanDamper1993} also use the maximum product of
preference values over all paths as one of their heuristics.

\cite{DamperEastmond1997} define a single heuristic strategy
called ``total product'' (TP'), which sums the products of arc
frequencies over \emph{all} paths that yield the same pronunciation.
\cite{DamperMarchand1998} consider a variation of the total
product rule, summing the products of arc frequencies over all
\emph{shortest} paths yielding the same pronunciation (TP).  Based on
the published accuracy scores, the results for this variation appear
to be somewhat better than for summing over all paths.
\cite{DamperMarchand1998} also consider another variant called
``weighted total product'' (WTP), where the product of arc frequencies
is divided by the product of segments lengths before summing, which
was found to yield a slight improvement.

In his method (SMPA) using unbounded overlap, \cite{Yvon1996}
replaced the shortest-paths primary heuristic with a one taking into
account the varying overlap.  His primary heuristic is equivalent to
choosing paths with the maximum average length of the segments.  The
secondary heuristic uses the frequency data to break ties as in
\cite{DedinaNusbaum1991} (we assume this means the sum of absolute
frequencies).

In their seminal article, Marchand and Damper (2000) proposed a method
for combining different strategies.  They use the shortest paths as
the primary heuristic but define several different secondary
strategies.  First, they rank the candidates using each of the
component strategies and award points to the candidates according to a
certain formula that gives the most points to the highest ranking
candidate and a decreasing number of points for each lower ranking
candidate.  The scores of the component strategies are then combined
by taking their sum or product.  The product rule was found to yield
slightly better results than the sum rule.

\subsection{Reimplementation of Best Published Algorithms}
\label{sec:reimpl}

A common problem well-known in the PbA literature is the difficulty of
replicating published results.  Differences in implementation details
typically yield different results for reimplemented algorithms.
Therefore, we have reimplemented the state of the art PbA algorithms
described in the literature to facilitate fair comparisons to our new
methods.

Several ideas have been proposed in the literature, but so far the
best published PbA results are obtained from segmentations with
overlap of one and the shortest-paths primary heuristic with various
secondary heuristics.  As a starting point, Table~\ref{tbl:bounds}
shows the lower and upper bounds for accuracy for the NETtalk corpus
given the primary heuristic of shortest paths.

In all published PbA works, the best performance has been obtain by
combinations of the following component strategies for evaluating the
candidate pronunciations as a secondary heuristic:
\begin{enumerate}
  \item Maximum product of arc frequencies (PF)
  \item Minimum standard deviation of arc lengths (SDPS)
  \item Maximum number of candidates with the same pronunciation (FSP)
  \item Minimum sum of the Hamming distances\footnote{The Hamming
    distance is the number of positions where two strings of equal
    length differ from each other.} of the candidate pronunciation
    from other candidates (NDS)
  \item Maximum smallest arc frequency of the path (WL)
  \item Maximum weighted product of arc frequencies (WPF): weighting
    by dividing each arc frequency by the number of different
    pronunciations for the letter sequence of the arc
  \item Maximum frequency of the first arc of the path (SF)
  \item Maximum frequency of the last arc of the path (SL)
  \item Maximum length of the longest arc of the path and in case of a
    tie, maximum frequency of the maximum frequency arc
    (SLN)\footnote{This definition appears to yield results closer to
      \cite{Polyakova2008} although their description seems to
      literally indicate maximum frequency of the longest arc as the
      secondary heuristic.}
  \item Maximum sum over letter positions of the number of other
    candidates having the same phoneme multiplied by the frequency of
    the arc containing the phoneme (SSPF)\footnote{It is unclear how
      to define the frequency of the containing arc for an overlap
      letter as it belongs to two arcs (or no arcs, depending on
      the definition).  We have included overlap letters twice, once for
      each arc. }
  \item Maximum sum of the geometric means of the products of arc
    frequencies over all candidates with the same pronunciation (PFSP)
\end{enumerate}
Strategies 1--5 were proposed by Marchand and Damper (2000) who
evaluated all $2^5-1=31$ combinations of the first five strategies;
the best result was obtained as a combination of all five.
Poly{\'a}kova and Bonafonte (2008, 2009) extended the set by
strategies 6--11 and evaluated all $2^{11}-1=2047$ combinations. The
best results were obtained by certain combinations not including all
strategies.  Table~\ref{tbl:bestpublished} shows the accuracy results
of our reimplementation of these methods.
\begin{table}
\centering
\begin{tabular}{l|cc|cc}
  & \multicolumn{2}{c|}{Text-to-Speech} &
    \multicolumn{2}{c}{Speech-to-Text} \\
     & Words (\%) & Phones (\%) & Words (\%) & Letters (\%)\\
  \hline
  Lower bound & 44.25 & (84.39) & 59.25 & (91.62) \\
  Upper bound & 85.42 & (96.68) & 89.76 & (98.08) 
\end{tabular}
\caption{\label{tbl:bounds}Lower and upper bounds for accuracy given
  the primary heuristic of shortest paths and \emph{overlap of one
    letter}.  The lower bound is obtained by unit score, which
  corresponds to tie-braking by random guessing.  The upper bound is
  given by the theoretically best possible secondary heuristic, which
  is obtained by minimizing the Levenshtein edit distance to the
  correct pronunciation (this implies that the correct pronunciation
  is chosen when available and otherwise the phoneme error is
  minimized).}
\end{table}

\begin{table}
\centering
\begin{tabular}{ll|cc|cc}
  \multicolumn{2}{c|}{Evaluation method}
  & \multicolumn{2}{c|}{Text-to-Speech} &
    \multicolumn{2}{c}{Speech-to-Text} \\
  Combination & Name & Words (\%) & Phones (\%) & Words (\%) & Letters (\%)\\
  \hline
  10000 & PF & 59.31 & (89.31) & \textbf{73.91} & \textbf{(94.78)} \\
  01000 & SDPS & 46.38 & (84.76) & 60.58 & (91.69) \\
  00100 & FSP & \textbf{62.46} & (89.90) & 72.41 & (94.42) \\
  00010 & NDS & 60.81 & \textbf{(89.96)} & 71.83 & (94.41) \\
  00001 & WL & 55.91 & (88.48) & 68.53 & (93.81) \\
  \hline
  10101 && 65.46 & (91.02) & \textbf{75.90} & (95.18) \\
  11111 && \textbf{65.62} & \textbf{(91.14)} & 75.52 & (95.10) \\
  \hline
  00000100000 & WPF & 59.25 & (89.18) & 74.06 & (94.82) \\
  00000010000 & SF & 58.28 & (88.84) & 68.00 & (93.67) \\
  00000001000 & SL & 51.16 & (86.84) & 69.58 & (93.80) \\
  00000000100 & SLN & 55.83 & (88.07) & 70.45 & (94.15) \\
  00000000010 & SSPF & 62.02 & (90.17) & 73.43 & (94.69) \\
  00000000001 & PFSP & \textbf{64.00} & \textbf{(90.57)} & \textbf{75.77} & \textbf{(95.21)} \\
  \hline
  $\cdots$ &&&& \\
  00101100011 && 65.55 & (91.12) & 76.09 & (95.24) \\
  00111111001 && 65.46 & (91.11) & 76.10 & (95.23) \\
  10111111001 && 64.94 & (90.98) & 76.12 & (95.23) \\
  00101001011 && 65.31 & (91.03) & 76.12 & (95.22) \\
  00101100001 && 65.35 & (90.99) & \textbf{76.19} & \textbf{(95.26)} \\
  $\cdots$ &&&& \\
  00111010011 && 66.05 & \textbf{(91.25)} & 75.71 & (95.15) \\
  01101000001 && 66.12 & (91.18) & 75.91 & (95.17) \\
  01100010001 && 66.13 & (91.22) & 75.59 & (95.12) \\
  00100010001 && 66.14 & (91.18) & 75.55 & (95.13) \\
  00101000001 && \textbf{66.14} & (91.20) & 76.00 & (95.22) \\
  \hline
  -- & TP & 61.76 & (89.96) & 74.91 & (94.99) \\
  -- & WTP & 62.02 & (90.04) & \textbf{74.97} & \textbf{(95.01)} \\
  -- & SMPA & \textbf{64.59} & \textbf{(90.55)} & 74.95 & (94.95)
\end{tabular}
\caption{\label{tbl:bestpublished}Accuracy scores of our
  reimplementation of the best published PbA methods.  First are shown
  the five strategies of Marchand and Damper (2000) followed by the
  best combinations for speech-to-text and text-to-speech.  Then are
  shown the six additional strategies of Poly\'akova and Bonafonte
  (2008,2009) followed by the best five combinations for
  speech-to-text and text-to-speech.  For comparison, the last three
  rows also show the performance of our reimplementation of the total
  product and weighted total product methods of Damper and Marchand
  (1998) and the SMPA method of Yvon (1996).  }
\end{table}

\subsection{Evaluation Methodology}

As in many previous works, we use the so called leave-one-out method
of evaluation.  That is, the accuracy of the algorithm for a given
word is evaluated by teaching all other words in the corpus and then
testing the generated pronunciation for the excluded word.  After
generating the pronunciation, we strip out all null phonemes from it
as well as from the correct pronunciation before evaluation.  This
improves performance as sometimes the algorithms generate the correct
pronunciation but with null phonemes at different places.\footnote{The
  stripping of null-phonemes is not done (or at least not reported) in
  the literature, but it seems like a more logical way of evaluating
  the algorithms as the null phonemes are only used for technical
  purposes.}

The \emph{word accuracy} of the algorithm is computed as the
proportion of correct generated pronunciations over all words in the
corpus.  If for a given word there are several candidate
pronunciations with the same score, we avoid arbitrary tie-braking by
computing the accuracy of the word as the number of correct candidates
with the best score divided by the total number of candidates with the
best score.

The \emph{phoneme accuracy} is computed by taking the Levenshtein edit
distance\footnote{The Levenshtein distance is the minimum number of
  edits (changing, adding, or removing one character) required to
  change one string to the other.} of the generated pronunciation from
the correct pronunciation and dividing it by the number of phonemes in
the correct pronunciation (not counting any silent null phonemes).  If
there are several candidates with the best score, then, as for word
accuracy, the phoneme accuracy is averaged over all of them.

When using overlapping segments, certain words are left without
pronunciation. This so called \emph{silence problem} occurs for
example for the word \emph{anecdote}, which does not get a
pronunciation as the substring \emph{cd} does not occur in any other
word.  We solve the silence problem when it occurs by allowing one
segment boundary without overlap at any position and then restricting
the resulting segmentations (e.g., by requiring minimal number of
segments) as usual.  This solution gives a pronunciation for all words
in the NETtalk corpus, but for the speech-to-text direction, leaves 12
words (0.06\%) without spelling.\footnote{We could generalize to allow
  two boundaries without overlap if we cannot get a pronunciation with
  one, but as we are mostly interested in the text-to-speech
  direction, we leave it for further work.}

In all our experiments, we use the NETtalk corpus, which is the most
widely used corpus in PbA work \citep{DamperEastmond1996,%
  DamperEastmond1997,DamperMarchand1998,DamperMarchand2000,%
  Yvon1996,DamperMarchand2006,SoonklangDamperMarchand2008,%
  Polyakova2008,Polyakova2009}.  It is based on a 20,009-word corpus
from the 1974 edition of Webster's Pocket Dictionary, and was used in
the NETtalk project in 1987 to train a neural network. The lexicon is
manually aligned by \cite{SejnowskiRosenberg} on a one-to-one
basis.  As in \citep{DamperMarchand2000}, we removed all homophones (or
homographs for the speech-to-text direction) and one-letter words from
the corpus, leaving 19,595 words (or 19,545 words for the
speech-to-text direction).\footnote{The numbers of words quoted by
  \cite{DamperMarchand2000} are one less than ours, which we
  assume is due to an error (last word listed twice) in the version of
  the lexicon cited by them. We use the lexicon in the version that is
  freely downloadable from
  \texttt{http://archive.ics.uci.edu/ml/datasets/Connectionist+Bench+(Nettalk+Corpus)}
} The speech-to-text direction is evaluated analogously to the
text-to-speech direction, we simply swap the words and their
pronunciations.

\section{A Probabilistic Approach to Pronunciation by Analogy}
\label{sec:probapproach}

In this section, we describe our simple probabilistic approach to PbA.
We present first a method based on the pronunciation lattice of
\cite{SullivanDamper1993}, which yields candidate pronunciations
with non-overlapping segments, as it is more straightforward to define
probabilities in this case.  In Section~\ref{sec:proboverlapping}, for
faster computation and better results, we generalize the method to
overlapping segments as used by among others
\cite{DedinaNusbaum1991} and \cite{DamperMarchand1998}.

\subsection{Training Data}

We assume that a lexical database is given, consisting of aligned
pairs $(x,y)=(x_1\dots x_l,y_1\dots y_l)$ of words and their
pronunciations.  Each word is assumed to be padded with the silent
space characters \emph{\#} at both ends.

We consider all substrings $(x_a\dots x_b,y_a\dots y_b)$, $1\le a\le
b\le l$, of every word-pronunciation pair $(x,y)$ in the corpus and
compute the frequencies of these substrings over the entire list of
words.  For each possible written substring $x$, the frequencies of
different corresponding pronunciations $y$ are then normalized as
\begin{equation}\label{eq:probest}
\hat p(y\mid x) = \frac{
  (\text{number of times $x$ is pronounced as $y$ in the training data})
}{
  (\text{number of times $x$ appears in the training data}) + 1
}
\end{equation}
to yield a probability distribution $\hat p(y\mid x)$.  This is the
training data for our PbA algorithm.  We add one to the denominator to
take into account the fact that it is possible that the segment $x$
has a novel pronunciation in the unseen word.  This is not necessary,
but it is probabilistically more consistent and slightly increases the
accuracy of our algorithm.

\subsection{Segmentations of a Word}

Given a written word $x$, we consider different \emph{segmentations}
$\len=(\len_1,\dots,\len_n)$ of the word, that is, lists of segment
lengths that sum to the length $l$ of the word.  For any given
segmentation $\len$ of the word $x$, we can compute a probability
distribution of possible pronunciations as
\begin{equation}
  p(y\mid x,\len) = \prod_{k=1}^n \hat p(y_k\mid x_k),
\end{equation}
where $x_k$ and $y_k$ denote the segments of the written word and its
pronunciation.

The choice of the segmentation greatly affects the resulting
distribution of its possible pronunciations.  For example, for the
string \emph{l}, the distribution of its pronunciations in the training
data might by given by the probability function\footnote{For
  simplicity, we do not include the novel pronunciation case of
  \eqref{eq:probest} in the examples.}
\[
\begin{tabular}{r@{ $\mapsto$ }l}
  \textipa{/l/} & $0.94$ \\
  \textipa{/ /} & $0.06$ \text(silent). 
\end{tabular}
\]
For the string \emph{y}, the distribution might be 
\[
\begin{tabular}{r@{ $\mapsto$ }l}
   \textipa{/I/} & $0.78$ \\
   \textipa{/aI/} & $0.22$.
\end{tabular}
\]
Thus, the string \emph{ly} segmented in two pieces would have the
following distribution of pronunciations:
\[
\begin{tabular}{r@{ $\mapsto$ }l}
  \textipa{/lI/} & $0.94 \cdot 0.78 = 0.7332$,\\
  \textipa{/lAi/} & $0.94 \cdot 0.22 = 0.2068$,\\
  \textipa{/ I/} & $0.06 \cdot 0.78 = 0.0468$,\\
  \textipa{/ Ai/} & $0.06 \cdot 0.22 = 0.0132$.
\end{tabular}
\]
However, if one considers \emph{ly} as one segment, the statistics
might be
\[
\begin{tabular}{r@{ $\mapsto$ }l}
   \textipa{/lI/} & $0.96$ \\
   \textipa{/lAI/} & $0.04$.
\end{tabular}
\]
Thus, the distributions for segmentations with a small number of
segments will generally be less spread out than the ones with a higher
number of segments (assuming, of course, that there are high order
correlations between the letters of the words of the language).

\subsection{Generating the Pronunciation}
\label{sec:genpron}

To generate the pronunciation, we consider all possible segmentations
$\len=(\len_1,\dots,\len_n)$ of the input word $x$
simultaneously. Each segmentation is considered as a potential
\emph{model} for having generated the true pronunciation of the
word. Hence, assuming a prior probability distribution $p(\len)$ over
all segmentations, we can average our pronunciation distribution over
all these models:
\begin{equation}\label{eq:modelaverage}
  p(y\mid x) = \sum_\len p(\len)p(y\mid x,\len).
\end{equation}
Assuming that one of the segmentations is the correct generating
model, the most likely true pronunciation for the written word $x$ can
then be output as
\begin{equation}\label{eq:mostlikely}
  \hat y := \argmax_y p(y\mid x),
\end{equation}
where $\argmax_y$ denotes the value of $y$ that maximizes the expression.

\subsection{The Choice of the Prior Distribution of Segmentations}

A good choice for $p(\len)$ is a distribution that assigns equal
probabilities to all feasible segmentations with the minimum number of
segments. Here a segmentation of the written word is considered
feasible if all of its segments can be found in the training data
(i.e., it corresponds to a path through the pronunciation lattice).
As we use a uniform prior distribution, the model average
\eqref{eq:modelaverage} is equivalent to simply summing (collating)
the estimated probabilities of all candidates corresponding to the
same pronunciation. 

The restriction to the minimum number of segments yields good
performance in practice since it yields maximally concentrated
probability distributions.  The same restriction is also applied in
the best-performing previously published PbA algorithms as described
in Sec.~\ref{sec:pronlattice}.  Another choice we have considered is a
uniform distribution over so called ``minimal segmentations'', that
is, segmentations such that you cannot get a feasible segmentation by
removing a segment boundary.  However, in our intitial experiments,
this did not improve performance.

\begin{table}
  \centering
  \begin{tabular}{l|cc|cc}
    & \multicolumn{2}{c|}{Text-to-Speech} &
    \multicolumn{2}{c}{Speech-to-Text} \\
    & Words (\%) & Phones (\%) & Words (\%) & Letters (\%)\\
    \hline
    Lower bound & 18.34 & (73.79) & 60.53 & (89.18) \\
    Upper bound & 91.18 & (98.32) & 94.12 & (99.02)
  \end{tabular}
  \caption{\label{tbl:boundsnonoverlapping}Lower and upper bounds for
    accuracy as in Table~\ref{tbl:bounds} given the primary heuristic
    of shortest paths for segmentations \emph{without overlap}.}
\end{table}

Table~\ref{tbl:boundsnonoverlapping} shows the upper and lower bounds
for accuracy for the NETtalk corpus given the restriction to minimum
number of segments for the prior distribution.  As expected, the
bounds are somewhat wider than for the more restrictive overlap of one
shown in Table~\ref{tbl:bounds}.

\subsection{Results and Discussion for Non-Overlapping Segments}

Our probabilistically justified evaluation method corresponds to a
\emph{collated} product of estimated probabilities, where
``estimated probabilities'' are defined by Eq.~\eqref{eq:probest} and
``collation'' means that the values of all candidates with the same
pronunciation are summed together.  Roughly speaking, this method is a
combination of certain ideas already suggested in the literature:
\begin{enumerate}
\item The product of arc frequencies has been used as the evaluation
  function by \cite{SullivanDamper1993} and
  \cite{DamperEastmond1997} with the justification of it being
  closer to a probabilistic definition than the sum as used by
  \cite{DedinaNusbaum1991} in their original program.
\item Collation has been used in the total product methods of
  \cite{DamperEastmond1997} and \cite{DamperMarchand1998} and
  it is also present in Strategy~3 of \cite{DamperMarchand2000},
  the best of the five component methods used therein, and in
  Strategy~11 of \cite{Polyakova2008}, the best of the 11 methods
  therein (see Sec.~\ref{sec:reimpl} for details).
\item Our estimated probabilities are similar to normalized
  frequencies as used by \cite{SullivanDamper1993} in one of their
  definitions of ``preference values'', but we add one to the
  denominator in order to better estimate the true probabilities.
\end{enumerate}
Our PROB method combines all three of the above and
Table~\ref{tbl:nonoverlapping} shows that this probabilistically
coherent method performs better than any other combination of
normalization type and collation or no collation.  It also performs
better than any of the five component strategies of
\cite{DamperMarchand2000} and better than any of the six new
strategies of \cite{Polyakova2008} except for Strategy~11 (we will
get to this strategy and why it works so well later in
Sec.~\ref{sec:strategy11}) and the generalization of our method for
overlapping segments will outperform even Strategy~11.  Thus, our
theoretically justified approach appears to work well also in
practice.

\begin{table}
\centering
\begin{tabular}{l|cc|cc}
  \multicolumn{1}{c}{} & \multicolumn{4}{c}{Text-to-Speech} \\
  \multicolumn{1}{c}{} \\
  & \multicolumn{2}{c|}{No collation} & \multicolumn{2}{c}{With collation} \\
  Evaluation method & Words (\%) & Phones (\%) & Words (\%) & Phones (\%) \\
  \hline
  Product of & & & & \\
  \multicolumn{1}{r|}{absolute frequencies}    & 47.37 & (85.72) & 53.03 & (87.26) \\
  \multicolumn{1}{r|}{normalized frequencies}  & 54.33 & (87.63) & 62.85 & (90.12) \\
  \multicolumn{1}{r|}{estimated probabilities} & 56.05 & (88.21) & \textbf{63.80} & \textbf{(90.51)} \\
  \multicolumn{1}{c}{} \\
  \multicolumn{1}{c}{} & \multicolumn{4}{c}{Speech-to-Text}\\
  \multicolumn{1}{c}{} \\
  & \multicolumn{2}{c|}{No collation} & \multicolumn{2}{c}{With collation} \\
  Evaluation method & Words (\%) & Letters (\%) & Words (\%) & Letters (\%)\\
  \hline
  Product of & & & & \\
  \multicolumn{1}{r|}{absolute frequencies}    & 69.83 & (93.92) & 71.96 & (94.36) \\
  \multicolumn{1}{r|}{normalized frequencies}  & 71.67 & (94.35) & 75.70 & (95.14) \\
  \multicolumn{1}{r|}{estimated probabilities} & 73.17 & (94.58) & \textbf{76.23} & \textbf{(95.22)} \\
\end{tabular}
\caption{\label{tbl:nonoverlapping}Results for non-overlapping
  segmentations with minimal number of segments.  The candidate
  pronunciations are evaluated by the product of absolute frequencies,
  normalized frequencies, or estimated probabilities as defined in
  Eq.~\eqref{eq:probest}.  Without collation, the candidate with the
  highest value is chosen; with collation, the values of all
  candidates with the same pronunciation are summed together first and
  the the pronunciation with the highest sum is chosen.  The method of
  collated estimated probabilities (results shown in bold) corresponds
  to our probabilistically justified method PROB and yields the
  unambiguously best results of those shown in the table.  
}
\end{table}

Indeed, normalization very clearly improves performance.  Why it does
so can be illustrated by pronouncing the word \emph{recent} using all
other words of the NETtalk corpus as the training set.  The algorithm
gives five different segmentations of two segments each and a total of
60 pronunciation candidates for the word. When absolute frequencies
are considered, the best candidate is the pronunciation rEkxnt
(\textipa{/rEk@nt/}), built from the segments \emph{\#rec} and
\emph{ent\#}. The absolute frequencies for these segments are 29 and
218, respectively, and their product is 6322. However, when
considering the normalized frequencies of the segment pronunciations,
the values are 0.392 and 0.729, showing that neither segment has a
particularly stable or unambiguous pronunciation.

When normalized frequencies are used, the best candidate pronunciation
is ris-Nt (\textipa{/ris-\s{n}t/}), formed from the segments
\emph{\#r} and \emph{ecent\#}, which have absolute frequencies of 903
and 2 but relative frequencies of 1.0 and 1.0, respectively. In other
words, the lexicon contains no words in which either segment is
pronounced in any other way, so the segment pronunciations are very
stable.

Collation also categorically improves performance in our experiments
and is justified by model averaging as discussed in
Sec.~\ref{sec:genpron}.

\section{A Probabilistic Approach for Overlapping Segments}
\label{sec:proboverlapping}

In many previous PbA algorithms, the pronunciations are generated
using overlapping segments. This avoids some of the potential problems
at the segment boundaries, in particular, it avoids certain impossible
pronunciations at the segment boundary such as two consecutive stop
consonants.  It also limits the number of possible segmentations and
therefore yields faster computation.  We can also generalize our
algorithm to use overlapping segments.

\subsection{Product Rule (PROD)}

Given a segmentation $\len=((a_1:b_1),\dots,(a_n:b_n))$, where the
segments $(a_k:b_k):=\{a_k,\dots,b_k\}$ cover all indices of the word
$x$, but can overlap each other arbitrarily, we can define
\begin{equation}
  p(y\mid x,\len) \propto \prod_{k=1}^n \hat p(y_{a_k:b_k}\mid x_{a_k:b_k})
\end{equation}
and use this definition with the same decision function
\eqref{eq:mostlikely} as with non-overlapping segments.  However, the
expression on the right is no longer a properly normalized
distribution as we require that the pronunciations of the overlapping
parts of the segments match each other.\footnote{The symbol $\propto$
  means that the right side should be divided by a normalization
  constant to yield a properly normalized distribution}  Thus, the
actual distribution resulting from this definition can be considered
as being generated by a procedure that samples the pronunciation for
each segment from their respective distributions and then outputs the
pronunciation for the full word only if the overlapping parts happen
to match each other.  This will bias the distribution of the
overlapping part towards the most common pronunciations.

\subsection{Conditional Probability Rules (CONDR, CONDL, CONDRL, CONDALL)}

A better justified definition that avoids the bias of the PROD rule
mentioned above is given by
\begin{equation}
p(y\mid x,\len) = \prod_{k=1}^n \hat p(y_{a_k:b_k}\mid x_{a_k:b_k},y_{a_1:b_1},\dots,y_{a_{k-1}:b_{k-1}}),
\end{equation}
where for any sequence $\mathbf{y} = y_{a_1:b_1},\dots,y_{a_n:b_n}$,
we define the estimated conditional probability expression used above
as
\begin{equation}
  \hat p(y_{a:b}\mid x_{a:b},\mathbf{y}) = 0
\end{equation}
if $\mathbf{y}$ does not agree with $y_{a:b}$ at some index, as
\begin{equation}
  \hat p(y_{a:b}\mid x_{a:b},\mathbf{y}) = 1
\end{equation}
if $\mathbf{y}$ fully covers and agrees with $y_{a:b}$, and as
\begin{equation}\label{eq:probest2}
\begin{split}
&\hat p(y_{a:b}\mid x_{a:b},\mathbf{y}) = \\
&\frac{\text{ (number of times $x_{a:b}$ is pronounced as
    $y_{a:b}$ in the training data) }}{\text{ (number of times
    $x_{a:b}$ appears with its pronunciation agreeing with
    $\mathbf{y}$)+1 }}
\end{split}
\end{equation}
otherwise.  

Thus, we sample the pronunciation for each segment from the
conditional distribution given the pronunciations of any overlapping
parts of the previous segments.  This is of course not symmetric, that
is, the resulting distribution depends on the order of the segments.
However, if one considers all orders (CONDALL) or just the
left-to-right and right-to-left orders (CONDRL) of the segments as
separate models, then symmetry is restored.  In addition to the
symmetric CONDALL and CONLR rules, we have also implemented and tested
the left-to-right order (CONDR) and right-to-left order (CONDL) rules.

\subsection{CONDF Rule}

A third way to handle the overlaps is as follows.  Let use denote by
$y_{\text{overlap}}$ the pronunciations of the overlapping letters of
a candidate pronunciation $y$.  Then, the CONDF rule is given by
\begin{equation}
  p(y\mid x,s) = \prod_{k=1}^n \hat p(y_{a_k:b_k}\mid x_{a_k:b_k},y_{\text{overlap}}).
\end{equation}
That is, we consider each segmentation and each configuration of the
overlap letters as a separate model and normalize the probabilities
given the pronunciations of the overlap letters.

\subsection{Example Calculations of Conditional Probabilities}

We shall use the word \emph{longevity} to demonstrate the different
normalization schemes used. Using a fixed one-letter overlap, the word
can be segmented into four different segmentations with a minimal
number of segments and these segmentations involve altogether ten
different segments:
\[
\begin{tabular}{l}
\#longe + evi + ity\# \\
\#longe + ev + vity\# \\
\#long + ge + evity\# \\
\#lon + nge + evity\# \\
\end{tabular}
\]
After removing the word itself from the corpus, the ten segments have
the following pronunciations and frequencies:
\[
\begin{tabular}{l|l}
\#longe & lcGg-: 1 \\
\#long  & lcG-: 4, lanJ: 2, lcGg: 1 \\
\#lon   & lcG: 5, lan: 2, lon: 1 \\
\hline
nge    & nJ-: 54, nJx: 18, Gg-: 12, nJE: 9, nJi: 9, G-{}-: 6, NJ-: 3, Ggx: 1, n-i: 1 \\
ge     & J-: 284, Jx: 105, JE: 80, Ji: 40, Z-: 26, g-: 19, -{}-: 16, gE: 11, -x: 8, \\
       & gx: 6, JI: 4, gA: 3, gi: 3, Ze: 2, -i: 1, Ja: 1, Za: 1, gI: 1, gY: 1, ge: 1 \\
ev     & Ev: 83, Iv: 50, iv: 36, -v: 24, xv: 15, Ef: 1 \\
evi    & ivi: 10, Evx: 8, IvA: 7, Ev-: 3, IvI: 3, ivA: 3, xvI: 3, -vI: 2, iv-: 2, \\
       & ivI: 2, -v-: 1, Evi: 1, IvY: 1, ivY: 1, ivy: 1, xvA: 1, xvY: 1 \\
\hline
evity\# & Evxti: 2 \\
vity\#  & vxti: 22 \\
ity\#   & xti: 421, Iti: 2 \\
\end{tabular}
\]

Despite the numerous different pronunciations that the middle segments
have, only few of them can occur in our pronunciation candidates since
the overlap between segments forces the pronunciations to overlap as
well. For the same reason the two-segment segmentation \#longe +
evity\# is not valid; the only pronunciations for the two segments do
not have a common phoneme overlapping on the letter \emph{e}.

The conditionality introduced by the overlap must also be taken into
consideration when normalizing the frequencies. There are up to four different
ways of normalizing each segment's frequency, since both the left and right
side of the segment might or might not have a pronunciation fixed for the
overlapping part. This in turn is dependent on the order of assigning
pronunciations for the segments in a segmentation; in a segmentation with $n$
segments there are $n!$ different orderings, which in our example case will mean
six different orderings for the word \emph{longevity}.

To demonstrate the different evaluation methods using conditionally
normalized frequencies, let us consider the segmentation \#lon + nge +
evity\# and its pronunciation lanJEvxti.  The first segment \#lon has
the beginning marker \# and will never overlap from its left
end. Similarly, the last segment evity\# with the ending marker will
never overlap from its right end.

The absolute frequencies of the segment pronunciations (lan, nJE and
Evxti) are 2, 9 and 2 respectively.  The unconditionally normalized
frequency for a segment pronunciation is the absolute frequency
divided by the sum of frequencies of all candidate pronunciations plus
one as shown in Eq.~\eqref{eq:probest}.

Conditional probabilities for the segments' pronunciations are derived
by dividing the pronunciation's absolute frequency by the sum (plus
one) of frequencies of all candidate pronunciations for the segment
that conform to the overlaps already fixed by the neighboring segments
as shown in Eq.~\eqref{eq:probest2}.

For our example segmentation and pronunciation, the normalized
frequencies and conditional probabilities are:
\[
\begin{tabular}{lccc}
  & lan & nJE & Evxti \\
\hline
unconditional normalization & $2/9$ & $9/114$ & $2/3$ \\
left overlap fixed &  & $9/92$ & $2/3$ \\
right overlap fixed & $2/4$ & $9/10$ &  \\
both overlaps fixed &  & $9/10$ &  \\
\end{tabular}
\]

In our conditional probability rules we define five evaluation methods that
utilize different orderings of the segments:

\begin{itemize}
\item CONDR: The pronunciation is constructed in a left-to-right order
  beginning with the leftmost segment. The first segment is normalized
  unconditionally and the construction continues to the right so that
  each consecutive segment's left overlap is fixed by the previous
  segment's pronunciation (black indicates a letter with a fixed
  pronunciation):
  \newcommand{\LL}{\multicolumn{1}{@{}|@{\hspace{2pt}}c@{\hspace{2pt}}}}
  \newcommand{\LR}{\multicolumn{1}{@{}|@{\hspace{2pt}}c@{\hspace{2pt}}|@{}}}
  \newcommand{\FLL}[1]{\multicolumn{1}{@{}|@{}c@{}}{\colorbox{black}{\color{white}#1}}}
  \newcommand{\FLR}[1]{\multicolumn{1}{@{}|@{}c@{}|@{}}{\colorbox{black}{\color{white}#1}}}
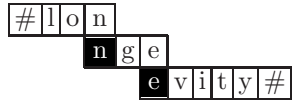
\[
\begin{tabular}{@{}c@{}c@{}c@{}c@{}c@{}c@{}c@{}c@{}c@{}c@{}c@{}}
\cline{1-4} 
\LL{\#} & \LL{l} & \LL{o} & \LR{n} & & &  &  &  &  & \\
\cline{1-6} 
 &  &  & \FLL{n} & \LL{g} & \LR{e} &  &  &  &  & \\
\cline{4-11} 
 &  & &  & & \FLL{e} & \LL{v} & \LL{i} & \LL{t} & \LL{y} & \LR{\#} \\
\cline{6-11} 
\end{tabular}
\]
Thus, the resulting estimated probability of our example pronunciation
is $(2/9)(9/92)(2/3) \approx 0.0145$.

\item CONDL: The pronunciation is constructed in a right-to-left order
  beginning with the rightmost segment. The first segment is
  normalized unconditionally and the construction continues to the
  left so that each consecutive segment's right overlap is fixed by
  the previous segment's pronunciation:
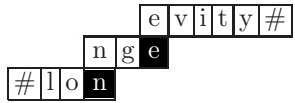
\[
\begin{tabular}{@{}c@{}c@{}c@{}c@{}c@{}c@{}c@{}c@{}c@{}c@{}c@{}}
\cline{6-11} 
 &  & &  & & \LL{e} &  \LL{v} & \LL{i} & \LL{t} & \LL{y} & \LR{\#} \\
\cline{4-11} 
 &  &  & \LL{n} & \LL{g} & \FLR{e} &  &  &  &  & \\
\cline{1-6} 
\LL{\#} & \LL{l} & \LL{o} & \FLR{n} & & &  &  &  &  & \\
\cline{1-4} 
\end{tabular}
\]
The resulting estimated probability of our example pronunciation is
$(2/3)(9/10)(2/4) = 0.3$.

\item CONDLR: This method takes the average of the results of the
  CONDR and CONDL methods to make the evaluation symmetrical, yielding
  an estimated probability of $\approx (0.0145+0.3)/2 \approx 0.157$

\item CONDALL: The pronunciation is assembled in all $n!$ orders and
  the average of their estimated probabilities is taken, yielding
  $\approx 0.022$. (Here the averaging corresponds to assuming a
  uniform prior distribution over the $n!=6$ possible models).

\item CONDF: This method first fixes the overlap letters and then
  normalizes each segment with both ends fixed (except at the
  beginning and end of the word):
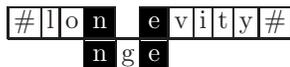
\[
\begin{tabular}{@{}c@{}c@{}c@{}c@{}c@{}c@{}c@{}c@{}c@{}c@{}c@{}}
\cline{1-4} \cline{6-11}
\LL{\#} & \LL{l} & \LL{o} & \FLR{n} & & \FLR{e} &  \LL{v} & \LL{i} & \LL{t} & \LL{y} & \LR{\#} \\
\hline
 &  &  & \FLL{n} & \LL{g} & \FLR{e} &  &  &  &  & \\
\cline{4-6} 
\end{tabular}
\]
In this case, the estimated probability of our example pronunciation
is $(2/4)(9/10)(2/3) = 0.3$.
\end{itemize}

Summed over all segmentations, the CONDL and CONDF methods in fact
yield the best score (probability) for the correct pronunciation while
the PROD method, the product of unconditional probabilities, does not.

\subsection{Results and Discussion for Overlapping Segments}

\begin{table}
\centering
\begin{tabular}{l|cc|cc}
  & \multicolumn{2}{c|}{Text-to-Speech} &
    \multicolumn{2}{c}{Speech-to-Text} \\
  Evaluation method & Words (\%) & Phones (\%) & Words (\%) & Letters (\%)\\
  \hline
  10000 & 59.31 & (89.31) & \textbf{73.91} & \textbf{(94.78)} \\
  00100 & \textbf{62.46} & (89.90) & 72.41 & (94.42) \\
  00010 & 60.81 & \textbf{(89.96)} & 71.83 & (94.41) \\
  \hline
  10101 & 65.46 & (91.02) & \textbf{75.90} & (95.18) \\
  11111 & \textbf{65.62} & \textbf{(91.14)} & 75.52 & (95.10) \\
  \hline
  00000000001 & \textbf{64.00} & \textbf{(90.57)} & \textbf{75.77} & \textbf{(95.21)} \\
  \hline
  00101100001 & 65.35 & (90.99) & \textbf{76.19} & \textbf{(95.26)} \\
  00111010011 & 66.05 & \textbf{(91.25)} & 75.71 & (95.15) \\
  00101000001 & \textbf{66.14} & (91.20) & 76.00 & (95.22) \\
  \hline
  PROB & 63.80 & (90.51) & 76.23 & (95.22) \\
  \hline
  PROD & 63.88 & (90.58) & 75.75 & (95.14) \\
  CONDF & \textbf{66.21} & \textbf{(91.13)} & 76.26 & (95.25) \\
  CONDR & 63.26 & (90.14) & 75.80 & (95.11) \\
  CONDL & 65.98 & (91.12) & 76.05 & (95.19) \\
  CONDRL & 65.50 & (90.98) & 76.30 & (95.27) \\
  CONDALL & 65.66 & (91.02) & \textbf{76.39} & \textbf{(95.29)} 
\end{tabular}
\caption{\label{tbl:results}The results of our probabilistic
  algorithms for overlapping segments compared to the best results of
  previously published algorithms from Table~\ref{tbl:bestpublished}.
  The probabilistically questionable PROD method is also shown for
  comparison.}
\end{table}

For overlapping segments, a strictly probabilistic evaluation is less
straightforward than for non-overlapping segments, because there is no
obvious way of defining the probability distribution of the
overlapping segments.  This is probably the reason why the approach of
\cite{SullivanDamper1993}, the one coming closest to a
probabilistic formulation (and using non-overlapping segments), has
been given up in later work, as overlapping segments are used
exclusively in the most recent work.  However, as our results will
show, our probabilistically justified evaluation rules are in fact
better than taking the direct product of segment probabilities.  Thus,
it appears that also for overlapping segments, our probabilistically
coherent theory works in practice.

Table~\ref{tbl:results} shows the results of our probabilistic
evaluation methods compared to the best previously published
algorithms.  The CONDF method performs better than any previously
published algorithm including all combinations of the 11 component
strategies of \cite{DamperMarchand2000} and
\cite{Polyakova2008}.  In fact, all of our probabilistically
justified left-right symmetric (i.e., excluding the asymmetric CONDR
and CONDL methods) methods perform better than any previously
published single strategy.

For the text-to-speech direction, the asymmetric CONDR and CONDL
strategies yield very different results, and interestingly, it is the
reverse right-to-left order of the segments that results in higher
accuracy.  For the speech-to-text direction, the difference between
CONDR and CONDL is much smaller, but still in the same direction.  The
symmetric CONDRL strategy, which includes both orders of the segments,
yields more consistent performance and is likely to maintain good
performance across different lexicons better.  The CONDALL strategy,
which includes all $n!$ orders of the segments offers a slight
improvement over CONDRL, and may be the best candidate for a general
method, but has a somewhat higher computational cost for long words.

\subsection{Strategy 11 of Poly{\'a}kova and Bonafonte (2008) and the Magic Root}
\label{sec:strategy11}

The only previously published single scoring strategy that works
better than PROB (our simplest probabilistic approach using
non-overlapping segments) is Strategy~11 of \cite{Polyakova2008}.
This is the collated geometric mean of the arc frequencies.  As we
have shown, collation categorically increases performance so it is not
surprising that this method works better than just the product of arc
frequencies.  However, what is surprising is that the geometric mean
(i.e., the $n$th root of the product of the $n$ arc frequencies)
actually works better than the product, contrary to any probabilistic
justification.

\begin{table}
  \centering
  \begin{tabular}{c|c@{\hspace{8pt}}c|c@{\hspace{8pt}}c|c@{\hspace{8pt}}c}
    \multicolumn{1}{c}{} & \multicolumn{6}{c}{Text-to-Speech}\\
    \multicolumn{1}{c}{} \\
    & \multicolumn{2}{c|}{Product of} & \multicolumn{2}{c|}{Product of} & \multicolumn{2}{c}{Product of} \\
    & \multicolumn{2}{c|}{absolute frequencies} & \multicolumn{2}{c|}{normalized frequencies} & \multicolumn{2}{c}{estimated probabilities} \\
    &&&&&&\\
    Root & Words (\%) & Phones (\%) & Words (\%) & Phones (\%) & Words (\%) & Phones (\%) \\
    \hline
    NC & 59.31 & (89.31) & 60.81 & (89.81) & 60.15 & (89.63) \\
    \hline
    $n$ & 64.00 & (90.57) & 65.45 & (91.00) & 65.56 & (91.04) \\
    \hline
    $1$ & 61.76 & (89.96) & 63.66 & (90.54) & 63.88 & (90.58) \\
    $2$ & 63.81 & (90.51) & 65.22 & (90.93) & 65.41 & (90.99) \\
    $3$ & 64.98 & (90.85) & 65.84 & (91.12) & 65.99 & (91.17) \\ 
    $4$ & 65.40 & (90.96) & 66.01 & (91.16) & 66.11 & (91.20) \\
    $5$ & 65.58 & (90.99) & \textbf{66.06} & \textbf{(91.18)} & \textbf{66.15} & \textbf{(91.21)} \\
    $6$ & 65.69 & (91.01) & 66.05 & (91.17) & 66.13 & (91.19) \\
    $7$ & 65.75 & (91.02) & 66.02 & (91.15) & 66.10 & (91.17) \\
    $8$ & \textbf{65.83} & \textbf{(91.04)} & 66.01 & (91.13) & 66.09 & (91.15) \\
    $9$ & 65.82 & (91.04) & 66.00 & (91.13) & 66.11 & (91.15) \\
    $10$ & 65.83 & (91.04) & 65.99 & (91.12) & 66.11 & (91.15) \\
    \multicolumn{1}{c}{}\\
    \multicolumn{1}{c}{}\\
    \multicolumn{1}{c}{} & \multicolumn{6}{c}{Speech-to-Text}\\
    \multicolumn{1}{c}{} \\
    & \multicolumn{2}{c|}{Product of} & \multicolumn{2}{c|}{Product of} & \multicolumn{2}{c}{Product of} \\
    & \multicolumn{2}{c|}{absolute frequencies} & \multicolumn{2}{c|}{normalized frequencies} & \multicolumn{2}{c}{estimated probabilities} \\
    &&&&&&\\
    Root & Words (\%) & Letters (\%) & Words (\%) & Letters (\%) & Words (\%) & Letters (\%) \\
    \hline
    NC & 73.91 & (94.78) & 74.37 & (94.92) & 74.51 & (94.92) \\
    \hline
    $n$ & 75.77 & (95.17) & 76.09 & (95.23) & 76.25 & (95.25) \\
    \hline
    $1$ & 74.91 & (94.99) & 75.50 & (95.12) & 75.76 & (95.14) \\
    $2$ & 75.68 & (95.17) & 75.98 & (95.21) & 76.12 & (95.23) \\
    $3$ & \textbf{75.99} & \textbf{(95.20)} & \textbf{76.14} & \textbf{(95.25)} & 76.26 & \textbf{(95.26)} \\
    $4$ & 75.97 & (95.19) & 76.12 & (95.24) & \textbf{76.27} & (95.26) \\
    $5$ & 75.98 & (95.19) & 76.04 & (95.22) & 76.18 & (95.24) \\
    $6$ & 75.94 & (95.18) & 75.96 & (95.20) & 76.11 & (95.22) \\
    $7$ & 75.91 & (95.17) & 75.96 & (95.19) & 76.09 & (95.21) \\
    $8$ & 75.91 & (95.16) & 75.91 & (95.17) & 76.05 & (95.19) \\
    $9$ & 75.88 & (95.15) & 75.88 & (95.16) & 76.03 & (95.18) \\
    $10$ & 75.87 & (95.14) & 75.86 & (95.16) & 75.98 & (95.17) \\
  \end{tabular}
  \caption{\label{tbl:pfsp} Accuracy obtained for the product of
    absolute frequencies, normalized frequencies, or estimated
    probabilities as the secondary evaluation function (as in
    Table~\ref{tbl:nonoverlapping}) for shortest paths with overlap of
    one.  The first row (NC) shows the results without collation.  The
    remaining rows show the results with collation and with the
    indicated root taken of the product before collation (without
    collation the root does not have any effect). Here $n$ is the
    number of segments in the candidate pronunciation and so the
    degree $n$ root of the product of absolute frequencies corresponds
    to the PFSP rule of Poly{\'a}kova and Bonafonte (2008).  However,
    these results indicate that a constant root is in fact better than
    the varying root.  Also, consistent with our other results,
    normalization increases accuracy. }
\end{table}

What strikes as odd in using the geometric mean is that the degree of
the root involved depends on the number of segments in the candidate
pronunciation.  This would perhaps be justifiable if the different
candidate pronunciations had different numbers of segments (so as to
bring the products on the same scale), but as the primary strategy
already limits the candidates to those with minimum number of
segments, it is difficult to justify why an effectively different rule
should be used depending on the number of segments.  In our
leave-one-out evaluation with the NETtalk corpus, the pronunciations
for most words can be constructed from two or three segments; thus,
the question arises if it would be better to instead use two or three
or some other constant as the degree of the root in the method.

Table~\ref{tbl:pfsp} shows the results of the product rule with
different constant roots and using the varying $kt$th root of
\cite{Polyakova2008}.  It turns out that a constant root is indeed
better and that the accuracy increases further by replacing the
absolute frequencies by the estimated probabilities as defined by
Eq.~\eqref{eq:probest}.  Still, the highest accuracy obtained is less
than that of CONDF, the best of our probabilistically justified methods
shown in Table~\ref{tbl:results}.

\begin{table}
\centering
\begin{tabular}{l|cc|cc}
  & \multicolumn{2}{c|}{Text-to-Speech} &
    \multicolumn{2}{c}{Speech-to-Text} \\
  Evaluation method & Words (\%) & Phones (\%) & Words (\%) & Letters (\%)\\
  \hline
  PROB & 63.80 & (90.51) & 76.23 & (95.22) \\
  PROB$^{1/2}$ & 65.25 & (90.98) & 76.66 & (95.36) \\
  PROB$^{1/3}$ & 65.60 & (91.14) & \textbf{76.83} & \textbf{(95.41)} \\
  PROB$^{1/4}$ & 65.81 & \textbf{(91.20)} & 76.74 & (95.38) \\
  PROB$^{1/5}$ & 65.87 & (91.19) & 76.69 & (95.37) \\
  PROB$^{1/6}$ & \textbf{65.92} & (91.19) & 76.65 & (95.37) \\
  PROB$^{1/7}$ & 65.89 & (91.19) & 76.58 & (95.35) \\
  PROB$^{1/8}$ & 65.88 & (91.17) & 76.51 & (95.33) \\
  PROB$^{1/9}$ & 65.88 & (91.16) & 76.51 & (95.33) \\
  PROB$^{1/10}$ & 65.88 & (91.15) & 76.52 & (95.33) \\
  \hline
  CONDF & 66.21 & (91.13) & \textbf{76.26} & \textbf{(95.25)} \\
  CONDF$^{1/2}$ & 66.29 & (91.15) & 76.17 & (95.21) \\
  CONDF$^{1/3}$ & 66.28 & (91.14) & 76.07 & (95.19) \\
  CONDF$^{1/4}$ & \textbf{66.31} & \textbf{(91.14)} & 76.02 & (95.17) \\
  CONDF$^{1/5}$ & 66.27 & (91.11) & 76.00 & (95.16) \\
  \hline
  CONDR & 63.26 & (90.14) & 75.80 & (95.11) \\
  CONDR$^{1/2}$ & 64.76 & (90.61) & 75.94 & (95.17) \\
  CONDR$^{1/3}$ & 65.20 & (90.77) & \textbf{76.00} & \textbf{(95.17)} \\
  CONDR$^{1/4}$ & \textbf{65.41} & \textbf{(90.84)} & 75.95 & (95.15) \\
  CONDR$^{1/5}$ & 65.46 & (90.85) & 75.93 & (95.14) \\
 \hline
  CONDL & 65.98 & (91.12) & 76.05 & (95.19) \\
  CONDL$^{1/2}$ & 66.52 & (91.31) & 76.20 & (95.23) \\
  CONDL$^{1/3}$ & \textbf{66.61} & \textbf{(91.33)} & \textbf{76.21} & \textbf{(95.24)} \\
  CONDL$^{1/4}$ & 66.54 & (91.31) & 76.17 & (95.23) \\
  CONDL$^{1/5}$ & 66.51 & (91.29) & 76.12 & (95.22) \\
  \hline
  CONDRL & 65.50 & (90.98) & 76.30 & (95.27) \\
  CONDRL$^{1/2}$ & 66.24 & (91.20) & \textbf{76.36} & \textbf{(95.27)} \\
  CONDRL$^{1/3}$ & \textbf{66.33} & \textbf{(91.23)} & 76.32 & (95.26) \\
  CONDRL$^{1/4}$ & 66.28 & (91.19) & 76.23 & (95.24) \\
  CONDRL$^{1/5}$ & 66.26 & (91.16) & 76.16 & (95.22) \\
\hline
  CONDALL & 65.66 & (91.02) & 76.38 & (95.29) \\
  CONDALL$^{1/2}$ & 66.30 & (91.23) & \textbf{76.39} & \textbf{(95.28)} \\
  CONDALL$^{1/3}$ & \textbf{66.39} & \textbf{(91.25)} & 76.36 & (95.27) \\
  CONDALL$^{1/4}$ & 66.35 & (91.22) & 76.26 & (95.24) \\
  CONDALL$^{1/5}$ & 66.31 & (91.18) & 76.17 & (95.21)
\end{tabular}
\caption{\label{tbl:root}Experiments applying the ``magic root''
  suggested by the good performance of the PFSP method of
  Poly{\'a}kova and Bonafonte (2008) to our probabilistically
  justified methods.  The degree of the applied root is shown as the inverse
  exponent of the method name.}
\end{table}

These results lead one to wonder if this ``magic root'' could be used
to improve the performance also in our probabilistically justified
methods.  This is indeed the case as shown by the results of
Table~\ref{tbl:root}.  Taking a root of the estimated probabilities
(before any summing) appears to improve the accuracy of all of our
probabilistic methods, except for the speech-to-text direction of the
CONDF method.  The optimum root appears to vary a bit depending on the
method and the lexicon, but based on these experiments, the cubic root
seems like a good compromise that works well for all methods.  The
CONDF method, which performed best, benefits the least from the root.
The asymmetric CONDL method obtains the best word accuracy (66.61\%)
of all methods when combined with the cubic root.

\section{Conclusion}

The current state of the art in pronunciation by analogy is well
summarized by the following passage from
\cite{DamperMarchand2006}:
\begin{quote}
  Automatic pronunciation of words from their spelling alone is a hard
  computational problem, especially for languages like English and
  French where there is only a partially consistent mapping from
  letters to sound. Currently, the best known approach uses an
  inferential process of analogy with other words listed in a
  dictionary of spellings and corresponding pronunciations. However,
  the process produces multiple candidate pronunciations and little or
  no theory exists to guide the choice among them.
\end{quote}
In this article, we have have presented a simple probabilistically
justified approach to choosing the best of several candidate
pronunciations.  Our principled approach outperforms all previously
published PbA algorithms (the best one by a small margin).  Thus, we
can obtain or exceed state of the art performance with a strictly
theoretically justified evaluation method without any ad hoc
modifications.

Our probabilistic approach differs from the more conventional methods by
a series of relatively small changes, each of which makes the method
closer to a strictly probabilistic formulation.  We have shown that
each of these changes improves performance.  Thus, even if one prefers
not to adopt our strictly probabilistic approach, one can still use
the following results yielded by our work in any PbA algorithms:
\begin{enumerate}
  \item Accuracy of a PbA algorithm can usually be improved by
    collation, that is, by summing together the values of all
    candidates with the same pronunciation.
  \item Instead of using absolute frequencies of the pronunciations,
    performance will typically increase by using normalized
    frequencies, or even better, estimated probabilities as defined by
    Eqs.~\eqref{eq:probest} and \eqref{eq:probest2}, instead.
  \item Some improvements of accuracy can be expected by replacing the
    frequencies or estimated probabilities of the segment
    pronunciations by their cubic root or some other root with degree
    $>1$. (This works in practice although we do not have a
    theoretical justification for it).
\end{enumerate}

Outside PbA literature, there are also other, statistically justified
approaches for text-to-speech conversion, see for example
\cite{Chen2003}, \cite{BisaniNey2008}, and
\cite{JiampojamarnKondrak2009}.  Based on reported experiments on
the NETtalk corpus, the state of the art accuracy in these methods is
actually better than in PbA; indeed, the best quoted word accuracy for
the NETtalk corpus is 69\% in \cite{BisaniNey2008}.  However,
these methods are far more complicated, both conceptually and
computationally than PbA algorithms; the programs run for several
hours and typically require setting aside a subset of words for tuning
certain global parameters of the algorithm.  The PbA algorithms remain
attractive due to their simplicity and comparatively good accuracy.

\section*{Acknowledgement}

This work was financially supported by the Academy of Finland grant
121855 and by the Tekes grant 40334/10.

\bibliographystyle{plainnat}

\begin{thebibliography}{16}
\providecommand{\natexlab}[1]{#1}
\providecommand{\url}[1]{\texttt{#1}}
\expandafter\ifx\csname urlstyle\endcsname\relax
  \providecommand{\doi}[1]{doi: #1}\else
  \providecommand{\doi}{doi: \begingroup \urlstyle{rm}\Url}\fi

\bibitem[Bisani and Ney(2008)]{BisaniNey2008}
Maximilian Bisani and Hermann Ney.
\newblock Joint-sequence models for grapheme-to-phoneme conversion.
\newblock \emph{Speech Communication}, 50:\penalty0 434--451, 2008.

\bibitem[Chen(2003)]{Chen2003}
Stanley~F. Chen.
\newblock Conditional and joint models for grapheme-to-phoneme conversion.
\newblock In \emph{Proceedings of European Conference on Speech Communication
  and Technology}, pages 2033--2036, Brighton, UK, 2003.

\bibitem[Damper and Eastmond(1997)]{DamperEastmond1997}
R.~I. Damper and J.~F.~G. Eastmond.
\newblock {Pronunciation by analogy: impact of implementational choices on
  performance}.
\newblock \emph{Language and Speech}, 40\penalty0 (1):\penalty0 1--23, 1997.

\bibitem[Damper and Marchand(2006)]{DamperMarchand2006}
R.~I. Damper and Y.~Marchand.
\newblock Information fusion approaches to the automatic pronunciation of print
  by analogy.
\newblock \emph{Information Fusion}, 7:\penalty0 207--230, 2006.

\bibitem[Damper and Marchand(1998)]{DamperMarchand1998}
R.I. Damper and Y.~Marchand.
\newblock Improving pronunciation by analogy for text-to-speech applications.
\newblock In \emph{Proceedings of 3rd European Speech Communication Association
  (ESCA)/COCOSDA International Workshop on Speech Synthesis}, 1998.

\bibitem[Damper and Eastmond(1996)]{DamperEastmond1996}
Robert~I. Damper and John F.~G. Eastmond.
\newblock Pronouncing text by analogy.
\newblock In \emph{Proceedings of the 16th conference on Computational
  linguistics}, pages 268--273, 1996.

\bibitem[Dedina and Nusbaum(1991)]{DedinaNusbaum1991}
Michael~J. Dedina and Howard~C. Nusbaum.
\newblock {PRONOUNCE}: a program for pronunciation by analogy.
\newblock \emph{Computer Speech and Language}, 5:\penalty0 55--64, 1991.

\bibitem[Glushko(1979)]{Glushko1979}
Robert~J. Glushko.
\newblock The organization and activation of orthographic knowledge in reading
  aloud.
\newblock \emph{Journal of Experimental Psychology: Human Perception and
  Performance}, 5:\penalty0 674--691, 1979.

\bibitem[Jiampojamarn and Kondrak(2009)]{JiampojamarnKondrak2009}
Sittichai Jiampojamarn and Grzegorz Kondrak.
\newblock Online discriminative training for grapheme-to-phoneme conversion.
\newblock In \emph{Proceedings of INTERSPEECH 2009}, pages 1303--1306, 2009.

\bibitem[Marchand and Damper(2000)]{DamperMarchand2000}
Yannick Marchand and Robert~I. Damper.
\newblock {A multistrategy approach to improving pronunciation by analogy}.
\newblock \emph{Computational Linguistics}, 26\penalty0 (2):\penalty0 195--219,
  2000.

\bibitem[Poly{\'a}kova and Bonafonte(2008)]{Polyakova2008}
Tatyana Poly{\'a}kova and Antonio Bonafonte.
\newblock {Further improvements to pronunciation by analogy}.
\newblock In \emph{Actas de las V Jornadas en Tecnolog\'{\i}a del Habla}, pages
  149--152, Bilbao, Spain, 2008.

\bibitem[Poly{\'a}kova and Bonafonte(2009)]{Polyakova2009}
Tatyana Poly{\'a}kova and Antonio Bonafonte.
\newblock {New strategies for pronunciation by analogy}.
\newblock In \emph{ICASSP '09}, 2009.

\bibitem[Sejnowski and Rosenberg(1987)]{SejnowskiRosenberg}
Terrence Sejnowski and Charles Rosenberg.
\newblock Parallel networks that learn to pronounce english text.
\newblock \emph{Complex Systems}, 1:\penalty0 145--168, 1987.

\bibitem[Soonklang et~al.(2008)Soonklang, Damper, and
  Marchand]{SoonklangDamperMarchand2008}
Tasanawan Soonklang, Robert~I. Damper, and Yannick Marchand.
\newblock Multilingual pronunciation by analogy.
\newblock \emph{Natural Language Engineering}, 1:\penalty0 1--21, 2008.

\bibitem[Sullivan and Damper(1993)]{SullivanDamper1993}
K.~P.~H. Sullivan and R.~I. Damper.
\newblock {Novel-word pronunciation: a cross-language study}.
\newblock \emph{Speech Communication}, 13\penalty0 (3-4):\penalty0 441--452,
  1993.

\bibitem[Yvon(1996)]{Yvon1996}
Fran{\c c}ois Yvon.
\newblock Grapheme-to-phoneme conversion using multiple unbounded overlapping
  chunks.
\newblock In \emph{Proceedings of the Conference on New Methods in Natural
  Language Processing, NeMLaP '96}, pages 218--228, Ankara, Turkey, 1996.

\end{thebibliography}

\end{document}